\documentclass[lettersize,journal]{IEEEtran}
\usepackage{tmi}
\usepackage{cite}
\usepackage{amsmath,amssymb,amsfonts}
\usepackage{algorithmic}
\usepackage{graphicx}
\usepackage{textcomp}
\usepackage{multirow}
\usepackage{braket}
\let\labelindent\relax
\usepackage[inline]{enumitem}
\usepackage[caption=false, font=footnotesize]{subfig}
\usepackage{array}
\usepackage[T1]{fontenc}
\usepackage[hyphens]{url}
\usepackage[hidelinks]{hyperref}
\hypersetup{breaklinks=true}
\newcommand{\system}{TransReg}
\begin{document}
\title{TransReg: Cross-transformer as auto-registration module for multi-view mammogram mass detection }
\author{Hoang C. Nguyen, Chi Phan, Hieu H. Pham, \IEEEmembership{Member, IEEE}
\thanks{Hoang C. Nguyen, Chi Phan, and Hieu H. Pham are with VinUni-Illinois Smart Health Center, VinUniversity, Hanoi, Vietnam.\\
E-mail: \{hoang.nc, 21chi.pth, hieu.ph@vinuni.edu.vn\}}
\thanks{Chi Phan, Hieu H. Pham is also with the College
of Engineering and Computer Science, VinUniversity, Hanoi, Vietnam.}}

\maketitle

\begin{abstract}
Screening mammography is the most widely used method for early breast cancer detection, significantly reducing mortality rates. The integration of information from multi-view mammograms enhances radiologists' confidence and diminishes false-positive rates since they can examine on dual-view of the same breast to cross-reference the existence and location of the lesion. Inspired by this, we present \system, a Computer-Aided Detection (CAD) system designed to exploit the relationship between craniocaudal (CC), and mediolateral oblique (MLO) views. The system includes cross-transformer to model the relationship between the region of interest (RoIs) extracted by siamese Faster RCNN network for mass detection problems. Our work is the first time cross-transformer has been integrated into an object detection framework to model the relation between ipsilateral views. Our experimental evaluation on DDSM and VinDr-Mammo datasets shows that our \system, equipped with SwinT as a feature extractor achieves state-of-the-art performance. Specifically, at the false positive rate per image at 0.5, \system\ using SwinT gets a recall at 83.3\%  for DDSM dataset and 79.7\% for VinDr-Mammo dataset. Furthermore, we conduct a comprehensive analysis to demonstrate that cross-transformer can function as an auto-registration module, aligning the masses in dual-view and utilizing this information to inform final predictions. It is a replication diagnostic workflow of expert radiologists
\end{abstract}


\section{Introduction}
\label{sec:introduction}
\begin{figure*}[htbp]
    \centering
    \includegraphics[width=\textwidth]{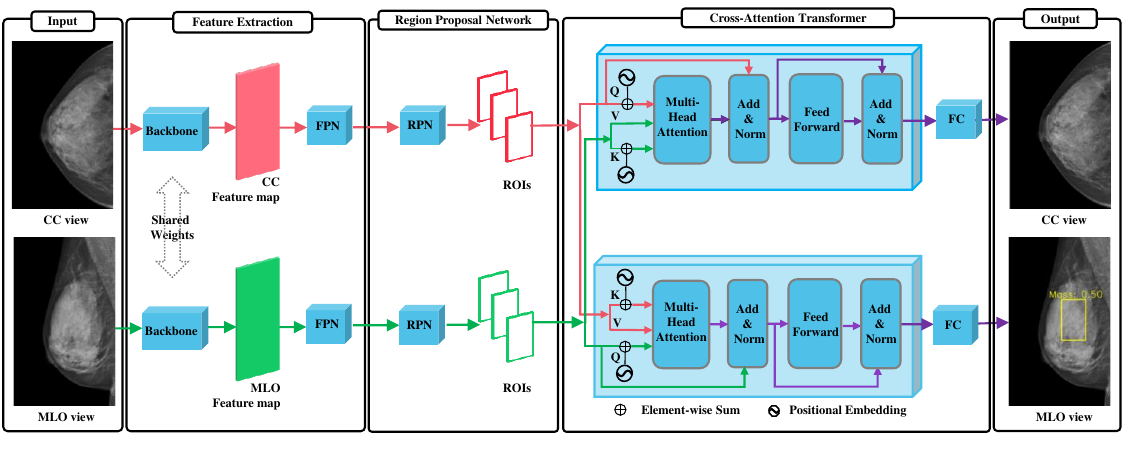}
    \caption{ \system\ architecture: The dual-views are encoded by shared weight Faster RCNN network to extract the region of interest (RoIs). The bidirectional cross-transformer network then leverages the cross-view information between RoIs from dual-views before making the final prediction. Positional encoding is added to the RoIs representation to include the spatial information   }
    \label{fig:main}
\end{figure*}

\IEEEPARstart{B}{reast} cancer is the most commonly diagnosed cancer worldwide accounting for about 2.3 million new cases and approximately 680,000 deaths in 2020 \cite{cancer}. The primary method for early detection of breast cancer is through screening mammography, which has been proven as a highly effective approach to reducing breast cancer mortality during its early stages \cite{cancer_reduce}. In a standard mammography procedure, four images are captured — two for each breast, consisting of a craniocaudal (CC) view and a mediolateral oblique (MLO) view. Radiologists frequently employ an ipsilateral analysis technique, where they examine dual views of the same breast to cross-reference and identify the presence and location of any abnormalities. This technique could significantly enhance diagnostic confidence and reduce the likelihood of false-positive results. Inspired by this approach, we develop  \system, a novel Computer-Aided Detection (CAD) system for mass detection. {\system} capitalizes on the intrinsic relationship between ipsilateral views by applying a cross-transformer \cite{transformer} to Regions of Interest (RoIs) extracted by a dual Faster RCNN network \cite{faster}. \\
In recent years, deep learning has made significant advancements in the field of medical imaging, leading to the development of computer-aided diagnosis (CAD) systems for mammograms \cite{Zhou2016MammogramCU}, \cite{Arevalo2015}, \cite{high}, \cite{lévy2016breast}, \cite{agarwal2019automatic},  \cite{ribli}, \cite{ballin}, \cite{9022155}, \cite{jung}. However, many existing CAD systems for mammography analysis predominantly rely on single-view images, overlooking the valuable insights that can be gained from the relationships between multiple views. Inspired by the workflow of radiologists, several deep neural network-based CAD systems have been designed to integrate information from dual-view to facilitate ipsilateral analysis. Nevertheless, to harness the full potential of multi-view images, the critical challenge lies in establishing correspondences between these views, which is known as image registration. Mammograms, due to their inherent properties, create considerable obstacles for image registration, including intensity changes and distortions induced by non-rigid deformations \cite{carneiroreg}. \\
One approach to tackle this registration challenge in mammogram analysis involves training the external models to align the lesson in dual ipsilateral views and subsequently utilizing that information for the final prediction \cite{perek}, \cite{YANyutong}. This method, however, requires an auxiliary task and cannot model the spatial relation between two different views. Moreover, as there is a lack of registration annotation between ipsilateral views, these works have to rely on assumptions to create positive pairs for training matching networks. Our proposed \system, instead, leverages cross-transformer to automatically register lesions and extract valuable features without the need for auxiliary tasks or manual registration annotations. Hence, {\system} could address the aforementioned limitations and offer a more effective solution for multi-view mammography analysis. \\
Other studies also explored the potential of implicitly learning the alignment between unregistered multiple mammography views. While some prior transformer-based methods, like the one proposed by Tulder et al. \cite{cross}, have shown promise in learning relations between ipsilateral views for feature maps, they are limited in dealing with high-resolution images due to the substantial memory and computational resources required for attention calculations on extensive feature maps. This has been shown to adversely impact the system performance \cite{high}. Furthermore, such methods are primarily designed for classification tasks and often lack a localization module. In contrast, our proposed system {\system} employs a cross-transformer applied to extracted RoIs, remaining conducive to the handling of high-resolution images. This allows {\system} to capture the relations between lesions, as represented by RoIs, from two distinct views and subsequently transform this information into valuable features for making accurate diagnostic decisions. Related to our research, CVR-RCNN \cite{ma2019crossview} uses Relation block \cite{Hu_2018_CVPR} to model the relation of the mass between CC and MLO views. Subsequently, Yang et al \cite{momminet} introduced IpsidualNet and later IpsidualNetv2 which recalculated the RoIs position based on the nipple position. Compared to transformers, these relation blocks lack crucial Feed-Forward Network (FFN) sublayers behind the attention sublayer, which plays an important role in capturing correlations and relationships among views. The integration of geometric information into the model also follows a complex and unconventional manner within Relation blocks. Our proposed model, instead, adopts a more straightforward approach by simply adding positional encoding to the RoIs's representation, akin to the methodology outlined in the original transformer paper \cite{transformer}. We also employ multi-head attention, as implemented in the transformer architecture, which enhances our model's capacity for feature learning compared to single-head attention used in prior systems. Another aspect worth considering is that previous studies did not conduct a comprehensive analysis to assess the model's capability in leveraging cross-view relations as radiologists employ in their diagnostic practice. 
Therefore, this paper aims to address those critical research gaps and introduce significant contributions to the field of mammography analysis in the following key aspects:  
\begin{itemize}
    \item We introduce \system, a novel multi-view detector using cross-transformers for ipsilateral views on mammograms. To the best of our knowledge, our work is the first time cross-transformers have been incorporated into an object detection framework for modeling the intricate relationships between CC and MLO views. This marks a significant step forward in optimizing multi-view mammography analysis, offering a fresh perspective on how cross-view information can be effectively harnessed.
    \item Our proposed \system\ outperforms all baseline and state-of-the-art methods on DDSM \cite{ddsm} and VinDr-Mammo \cite{vindr} datasets in Free-Response Operating Characteristic (FROC) mass detection. Remarkably, when employing SwinT as the feature extractor network, our dual-view approach surpasses even tri-view state-of-the-art models  \cite{YANG2021102204}, \cite{agrcnn}. Our codes are made publicly available at [\url{https://github.com/levi3001/multiview-mamo}]
    \item We conduct extensive experiments to demonstrate that \system, using cross-transformer, have the ability to register masses in CC and MLO views and effectively utilize the cross-view information to generate diagnostic predictions automatically. These experiments thereby thoroughly analyze the capacity of our model to replicate the natural diagnostic workflow followed by expert radiologists, which serves as a reliable testament to the practical utility and the alignment of a CAD system to clinical practices. 
    
\end{itemize}

The rest of our paper is organized as follows. Related works are reviewed in section \ref{sec:related}. In section \ref{sec:main}, we formulate the problem and describe the proposed method and model architecture. In section \ref{sec:exp}, we provide details on our experiment setup and result. Section \ref{sec:anal} analyze \system\ ability to leverage dual-view information. Finally, we conclude the paper in section \ref{sec:conclude}, discussing its limitations and outlining possible future research directions.

\section{Related Works}
\label{sec:related}
\begin{figure*}[htbp]
     \centering
     \begin{tabular}{c c c c }

         \includegraphics[width=0.25\textwidth]{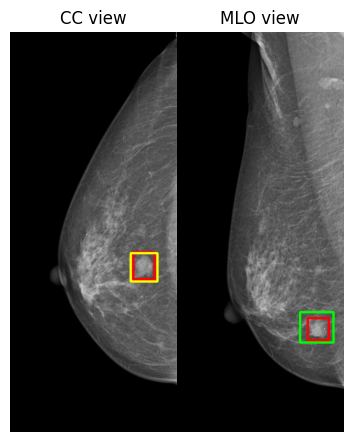}
     & \hspace{-6.5mm}

         \includegraphics[width=0.25\textwidth]{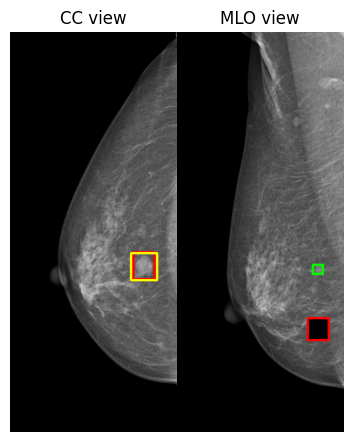}  
     &

         \includegraphics[width=0.25\textwidth]{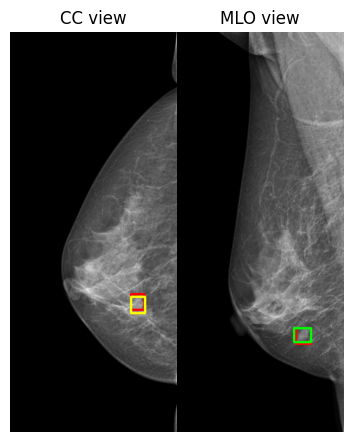}
     & \hspace{-6.5mm}

         \includegraphics[width=0.25\textwidth]{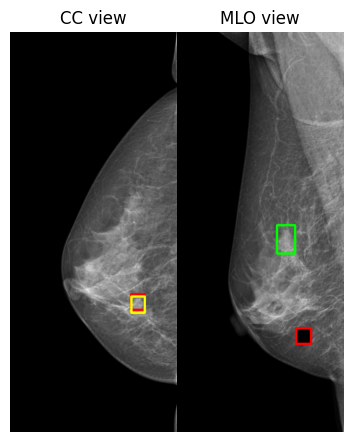}
    \vspace{-3mm}
    \\ 

         \includegraphics[width=0.25\textwidth]{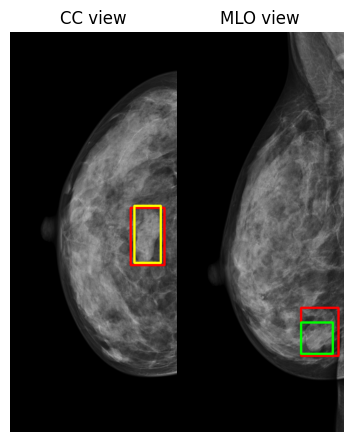}
     &  \hspace{-6.5mm}

         \includegraphics[width=0.25\textwidth]{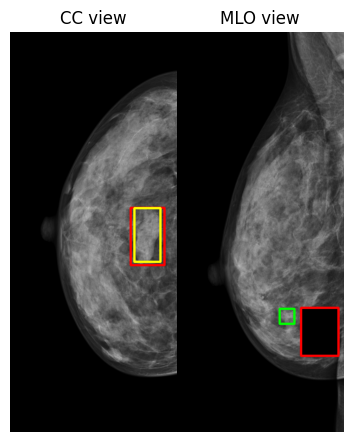}  

     & 

         \includegraphics[width=0.25\textwidth]{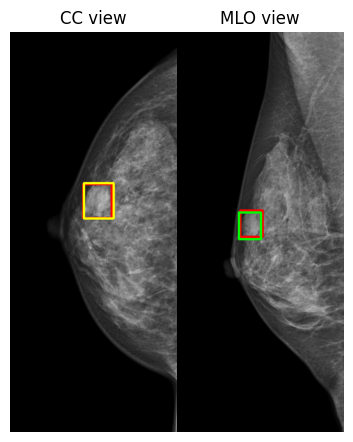}

     & \hspace{-6.5mm}

         \includegraphics[width=0.25\textwidth]{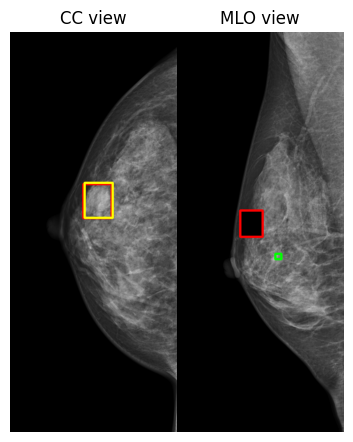}
 
    \vspace{-3mm}
    \\ 

         \includegraphics[width=0.25\textwidth]{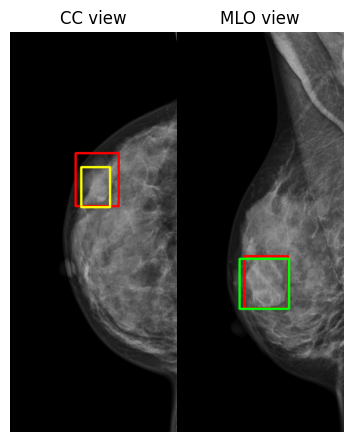}

     & \hspace{-6.5mm}

         \includegraphics[width=0.25\textwidth]{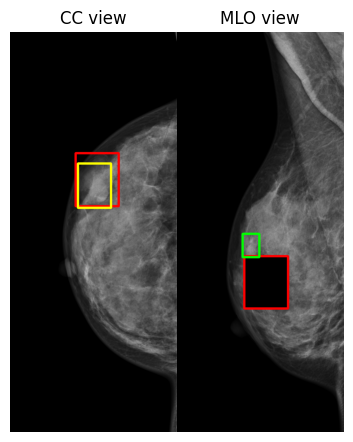}  

     &

         \includegraphics[width=0.25\textwidth]{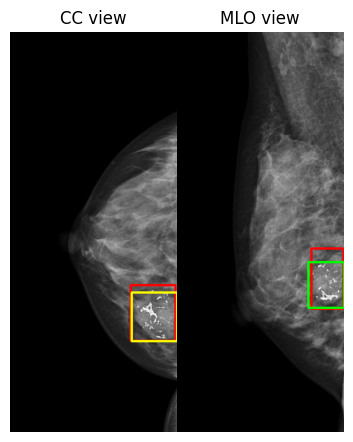}

    & \hspace{-6.5mm}

         \includegraphics[width=0.25\textwidth]{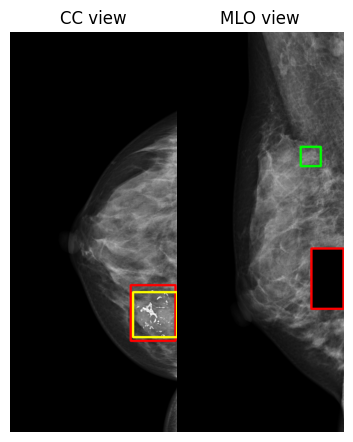}

    \end{tabular}
        \caption{Vizualization of the highest relevance score box in MLO view corresponding to the prediction box in CC view for image from VinDr-Mammo dataset and masked synthesis dataset.Column 1 and 3 refer to the views from original datasets and column 2 and 4 refer to the corresponding views in synthesis datasets. The red boxes are ground truth annotation, the yellow boxes are the prediction boxes of the model on the CC view, and green boxes are the proposals on MLO view have the highest relevance score in section \ref{subsec:viz} corresponding to the predictions}
        \label{fig:viz}
\end{figure*}


\subsection{Breast Cancer Detection on Mammograms:}
Detecting breast cancer on mammograms is a crucial task that has seen significant progress over the past few decades, especially through Computer-Aided Detection (CAD) systems. Early CADs addressed this problem by using handcrafted features \cite{Renato_Campanini_2004}, \cite{4214877}, \cite{sampat}, which often resulted in weak representations and a relatively high false positive rate. In recent years, with the development of Deep Learning techniques in Computer Vision (CV) and Natural Language Processing (NLP), Deep Neural Network (DNN) models have have been increasingly applied for classification \cite{carneiromicai}, \cite{Zhou2016MammogramCU}, \cite{Arevalo2015}, \cite{high}, \cite{cross}, \cite{lévy2016breast} and localization \cite{teare}, \cite{agarwal2019automatic}, \cite{ribli}, \cite{ballin}, \cite{Famouri}, \cite{Dhungel15}, \cite{9022155}, \cite{jung} of mass or calcification in mammograms, delivering a substantial enhancement in performance. A common practice in breast cancer detection using deep models is to employ multitask learning, which initially localizes the lesions within mammograms and subsequently employs the predicted lesion locations along with the original images to make final diagnostic decisions \cite{carneirotmi} \cite{nanwu} \cite{momminet} \cite{YANG2021102204} \cite{agrcnn}. Various approaches have been explored to localize the lesion on mammograms. Some have adopted a patch-level classification strategy, which involves dividing the mammogram image into smaller patches and training classification models on these patches before aggregating them to localize the abnormality \cite{agarwal2019automatic}, \cite{nanwu}.  Another common approach is to utilize modern region base detectors such as Faster RCNN \cite{faster}, Mask-RCNN \cite{maskrcnn}, RetinaNet \cite{retina}, Yolov3 \cite{yolov3} to train in an end-to-end manner ( \cite{ribli}, \cite{ballin}, \cite{Famouri}, \cite{Dhungel15}, \cite{9022155}, \cite{jung}, \cite{YANyutong}). However, despite these advancements, there remains a significant gap in exploiting the complementary nature of multi-view mammograms to improve breast cancer diagnosis.
\subsection{Multi-view mammogram analysis}
Numerous ongoing efforts have been made to harness the multi-view information in breast cancer diagnosis. One relatively straightforward yet effective approach is to encode the view and concatenate the representation \cite{carneirotmi}, \cite{nanwu}. Carneiro et al. \cite{carneirotmi} analyze the strategy and stage (early or late) for merging the representation. This method, however, is not specifically designed to model the relation between bilateral or ipsilateral view, and the concatenating feature may not be well-suited for object detection frameworks. Liu et al. \cite{CBN} introduced a contrasted bilateral network (CBN) to leverage the information of bilateral view. Perek et al \cite{perek} and Yan et al. \cite{YANyutong} train matching networks between extracted RoIs from ipsilateral pairs to register the lesions between CC/MLO views of the same breast. Yan et al. \cite{YANyutong} adopts multi-task learning to train patch matching and classification networks jointly for robust generic feature extraction. This approach requires explicitly training a new model without registration annotations for aligning ipsilateral views. Additionally, the geometry information is incorporated by the design.  CVR-RCNN \cite{ma2019crossview} use Relation block \cite{Hu_2018_CVPR} to model the relationships of the masses between both views. Momminet v1, v2 \cite{momminet}, \cite{YANG2021102204} utilize ipsilateral analysis based on CVR-RCNN and also bilateral analysis with a total of three views. Relation Network used in these studies integrates spatial information in a complex and unconventional way and the network itself lacks a Feed Forward Network compared with Transformer. Furthermore, these works do not establish the necessary experiments to prove that their model can work as intuition. Liu et al. \cite{agrcnn} introduced AG-RCNN using a graph convolution network to process tri-view mammograms. This work, however, requires many complex pre-process and post-process steps to utilize the graph convolution network. In this paper, we introduce {\system}, a novel approach in multi-view mammogram analysis that is capable of extracting valuable features without auxiliary tasks or manual registration annotations by cross-transformer. For the first time, a cross-transformer module has been incorporated into a mass detection model to leverage the cross-view information of mammogram images and has been studied extensively to demonstrate its effectiveness.

\section{Proposed Approach} 
\label{sec:main}
\subsection{Problem formulation}
Given a pair of CC/MLO view $\mathbf{x}=(\mathbf{x}_{\mathrm{CC}}, \mathbf{x}_{\mathrm{MLO}})$, we aim to generate predictions $\mathbf{y} =\set{\mathbf{y}_{\mathrm{CC}}, \mathbf{y}_{\mathrm{MLO}}}$ where $\mathbf{y}_{k} = (\mathbf{c}_{k}, \mathbf{b}_{k})$, $k \in \set{\mathrm{CC, MLO}}$. Here $\mathbf{b}_{k} \in \mathbb{R}^{4 \times M}$ represents the bounding box of the finding mass, $\mathbf{c}_{k} \in (0,1)^{M}$ stands for confident score of the prediction and $M$ is the number of predicted masses within view. 
\subsection{Baseline model}
For single view detector baseline, our goal is to determine the function $f$ such that $\mathbf{y}_{k}= f(\mathbf{x}_{k})$ for each view $k \in \set{\mathrm{CC, MLO}}$ separately. For modern detectors, $f$ has form $f= h \circ g$, where $h$ is responsible for extracting the list of the region of interests (RoIs). We denote $h(\mathbf{x}_{k}) = \mathbf{p}_{k}$ with $\mathbf{p}_{k} \in \mathbb{R}^{P \times d}$ representing the list of RoIs, $P$ and $d$ denoting the number of RoIs and the number of dimensions respectively. Function $g$ then use these RoIs to make final prediction, resulting in $\mathbf{y}_{k}= g(\mathbf{p}_{k})$. In this work, we use Faster RCNN architecture for single-view baseline. These models consist of an encoder (also referred to as the backbone) responsible for feature map learning. Subsequently, a Region Proposal Network (RPN) is employed to extract Regions of Interest (RoIs), which are then used for both lesion classification and localization. Following the setting suggested in Yang et al \cite{momminet}, we incorporate Feature pyramidal network (FPN) \cite{fpn}, focal loss \cite{retina}, and Distance IOU (DIOU) \cite{DIOU} loss to the framework. We use Resnet50 \cite{resnet} and SwinT \cite{swin} as feature extractors.
\subsection{Multiview detector overview}
For multi-view detector, the model takes both views as input, i.e., $\mathbf{y}_{\mathrm{CC}}, \mathbf{y}_{\mathrm{MLO}} = f(\mathbf{x}_{\mathrm{CC}}, \mathbf{x}_{\mathrm{MLO}})$. Our proposed \system\ models first extract the RoIs from ipsilateral view (CC and MLO) using dual Faster RCNN and combine them with cross-view transformer block as illustrated in Fig \ref{fig:main}. Motivated by a line of study using Siamese structure for dual-view mammograms \cite{perek}, \cite{ma2019crossview}, \cite{YANG2021102204} two branches Faster RCNN network shared weight to extract features from two views in the same way and to reduce the memory and computation resources. We formulate it as $\mathbf{p}_{\mathrm{CC}}, \mathbf{p}_{\mathrm{MLO}} = h(\mathbf{x}_{\mathrm{CC}}, \mathbf{x}_{\mathrm{MLO}}) = h'(\mathbf{x}_{\mathrm{CC}}), h'(\mathbf{x}_{\mathrm{MLO}})$. In contrast to the single-view detector, $g$ combines the RoIs from dual-view to make the final prediction $\mathbf{y}_{\mathrm{CC}}, \mathbf{y}_{\mathrm{MLO}}= g(\mathbf{p}_{\mathrm{CC}}, \mathbf{p}_{\mathrm{MLO}})$. In $g$, a cross-transformer module is used to encode the relation between RoIs, which aligns with the area containing abnormality in the image, extracted from Faster RCNN. This enables \system\ to effectively utilize the information between both views. Similar to our baseline, we incorporate FPN, focal loss, DIOU loss, and employ Resnet50 \cite{resnet} and SwinT \cite{swin} as backbone network.

\subsection{Cross-transformer}
We utilize the bidirectional cross-transformer used in transformer decoder in Vaswani et al. \cite{transformer}. For each direction, we consider RoIs list from the main view ($\mathbf{p}_{m}$) and use the corresponding ipsilateral view as an auxiliary view ($\mathbf{p}_{a}$) to compute co-attention. Each cross-transformer block consists of two key components: multi-head co-attention (MCA) and feed-forward network (FFN) combined with residual connection as in equation \eqref{cross}. We use post norm setting so Layer norm (LN) is applied after each component. 
\begin{equation} \label{cross}
\begin{aligned}
    \mathbf{p}_{m} = \mathrm{LN}(\mathbf{p}_{m} + \mathrm{MCA}(\mathbf{p}_{m},\mathbf{p}_{a})) \\
    \mathbf{p}_{m} = \mathrm{LN}(\mathbf{p}_{m}+ \mathrm{FFN}(\mathbf{p}_{m}))
\end{aligned}
\end{equation}

Equation \eqref{attn} computes the single-head attention, for multi-head attention, please refer to the original paper \cite{transformer}. We get the query $Q$ as linear projection (Prj) of $\mathbf{p}_{m}$- list of RoIs for the main view and key and value $K$ and $V$ as linear projection of $\mathbf{p}_{a}$ - list of RoIs for the auxiliary view. To encode the spatial information, unlike Ma et al. \cite{ma2019crossview}, we simply adopt 2D positional encoding (Pos) as in Vaswani et al. \cite{transformer}. This positional encoding uses the center of the proposal aligned with the RoI as input. The cross-attention module effectively serves as an auto-registration module where we compute the similarity in both context and position of the RoIs between main and auxiliary views. This similarity is then utilized as a weight to combine the RoIs from the auxiliary view to the main view

\begin{equation} \label{attn}
   \mathrm{Attention}(Q, K, V) = \mathrm{softmax}(\frac{QK^T}{\sqrt{d}})V
\end{equation}
\begin{equation}
\begin{aligned}
    Q = \mathrm{Prj_{q}}(\mathbf{p}_{m})+ \mathrm{E_{Pos}}(p_{m})\\
    K = \mathrm{Prj_{k}}(\mathbf{p}_{a})+ \mathrm{E_{Pos}}(\mathbf{p}_{a})\\
    V= \mathrm{Prj_{v}}(\mathbf{p}_{a})
\end{aligned}
\end{equation}
\section{Experiments} 
\label{sec:exp}
\subsection{Datasets and Experimental Settings} 

We perform experiments on both public datasets DDSM \cite{ddsm} and  VinDr-Mammo \cite{vindr}. Details are provided below.\\
\textbf{DDSM Dataset}: Digital Database for Screening Mammography (DDSM) was collected by the University of South Florida. The dataset is digitalized from screen-film mammography (SFM) including 2620 study cases. Each study case contains four views (left and right craniocaudal (CC) and mediolateral oblique (MLO)). The dataset provides bounding boxes and mask annotation for localize the finding lesions for both mass and calcification. Additionally, each lesion is categorized as malignant or benign, and the cancer studies have historical proof. To align with previous research \cite{momminet} \cite{YANG2021102204}, \cite{CBN} we split the dataset study cases into train/validate/test with the rate $80\% / 10\% / 10\%$. \\
\textbf{VinDr-Mammo Dataset}: 
VinDr-Mammo Dataset is full-field digital mammography (FFDM) dataset that consists of 20000 images derive from 5,000 study cases. These study cases were randomly sampled from the pool of all mammography examinations taken between 2018 and 2020 via the Picture Archiving and Communication System (PACS) from two Vietnam's hospital namely Hanoi Medical University
Hospital (HMUH–https://hmu.edu.vn/) and Hospital 108 (H108–https://www.benhvien108.vn/home.htm). Therefore, the dataset represents the real distribution of patients observed in these hospitals. The dataset offers bounding box annotations for localization lesions of various classes such as mass, calcification, asymmetries, architectural distortion,  suspicious lymph node, skin thickening, skin retraction, and nipple retraction. It includes BI-RADS assessment for mass, calcification, asymmetries, and architectural distortion. The dataset has been already splited into train/test with 1000 exams for testing and the rest for training.
\subsection{Evaluation metrics} 
To evaluate the effectiveness of the proposed method, we use FROC (free response receiver operating characteristic) which calculate the recall at different false positive per image  (FPPI) as the evaluation metric. A detected mass region is recalled if its IOU with growth truth is greater than 0.2.

\subsection{Implementation details}
As suggestion by Geras et al. \cite{high}, we use high-resolution images to train the models. For VinDr-mammo datasets, we resize the image to $500 \times 1200$ and for DDSM dataset, since the mammogram has low quality, we use a larger image size at $1000 \times 1500$. In VinDr-Mammo dataset, each image has large background area so we crop the breast using histograms. We use Adam optimizer \cite{adam} with a learning rate of 1e-5 for all of our models. Augmentation techniques such as horizontal and vertical flips, Gaussian noise, and box scaling \cite{midl} are applied. For post-processing the bounding box, we use the common strategy proposed in Ribli et al. \cite{ribli}, to fix the nms threshold to 0.1 
\subsection{Experiment result}
DDSM Dataset Evaluation: We evaluate \system\ performance on DDSM dataset and compare it with several baselines and recent state-of-the-art methods. The results are summarized in Table \ref{table:rddsm}. For the dual-view model, since our model is designed to handle ipsilateral view, we only consider the work working with these dual views such as CVR-RCNN, IspidualNetv1, v2, and BG-RCNN. We also compare \system\ performance with frameworks utilizing three views (Momminet v1, v2, and AG-RCNN). Except for BG-RCNN which involves extensive preprocessing steps, our models outperform all single and dual models when using Resnet 50 as feature extractors. Furthermore, \system\ demonstrated competitive performance with Momminet v1 which uses tri-view. Remarkably, when SwinT was used as the feature extractor, \system\ outperformed all single, dual, and tri-view methods. \\
VinDr-Mammo Dataset Evaluation: To show the robustness of our proposed method, we conduct an evaluation on VinDr-Mammo dataset. The result presented in Table \ref{table:rvin}, demonstrates \system\ consistently outperformed the corresponding baselines across all False Positive Per Image (FPPI) levels.

\begin{table*}[htbp]
\caption{FROC analysis for DDSM dataset}
\label{table:rddsm}
\normalsize
\centering

\begin{tabular}{lllccc}
\hline
View & & Data split &\multicolumn{3}{c}{Recall @ FPPI} \\ 
 & &   & R@0.5 & R@1 & R@2  \\ \hline
\multirow[t]{5}{*}{\textit{Single}} &Campanini et at. \cite{Renato_Campanini_2004} &1400/\_/512 & $\sim$ 0.54 & $\sim$ 0.74 & $\sim$ 0.86   \\   
&Sampat et at. \cite{sampat}  &349/150/100 & N/A & $\sim$ 0.803 & N/A  \\   
&Faster RCNN HRnet FPN Focal DIOU \cite{YANG2021102204} & 80\%/10\%/10\%   & 0.76 & 0.82 & 0.88   \\  
&Faster RCNN Resnet50 FPN Focal DIOU (ours) & 80\%/10\%/10\%  &0.743 &	0.833 &	0.865  \\  
&Faster RCNN Swint FPN Focal DIOU (ours)  & 80\%/10\%/10\% & 0.780	 & 0.861	& 0.918  \\ \hline
\multirow[t]{4}{*}{\textit{Dual}}& CVR RCNN \cite{ma2019crossview}& 410/\_/102 & N/A & N/A & $\sim$ 0.88     \\ 
& IpsiDualNetv1 (Resnet50) \cite{momminet} & 80\%/10\%/10\% & 0.764 & 0.828 & 0.879     \\ 
& \system\ (Resnet50) (ours) & 80\%/10\%/10\%  & 0.784 &	0.845 &	0.894   \\
&BG-RCNN \cite{bgrcnn} & 70\%/10\%/20\% & 0.795 & 0.866 & 0.918    \\ 
& IpsiDualNetv2 (HRnet) \cite{YANG2021102204} & 80\%/10\%/10\% &  0.81 & 0.84 & 0.89     \\ 
& \textbf{\system\ (Swint) (ours)} & 80\%/10\%/10\%  & \textbf{0.833} &	\textbf{0.898}	& \textbf{0.931}  \\ \hline
\multirow[t]{3}{*}{\textit{Tri}}& Momminet v1 (Resnet50) \cite{momminet} & 80\%/10\%/10\% &  0.802 & 0.849 & 0.892   \\ 
& Momminet v2 (HRnet)  \cite{YANG2021102204} & 80\%/10\%/10\% &    0.831 & 0.850 & 0.898    \\ 
& AG-RCNN \cite{agrcnn} & 70\%/10\%/20\% &    0.820 & 0.890 & 0.921  \\\hline
\end{tabular}

\end{table*}

\begin{table*}[htbp]
\normalsize
\centering
\caption{FROC analysis for VinDr-mammo dataset.}
\begin{tabular}{llllll}
\hline
View &   &\multicolumn{3}{c}{Recall @ FPPI} \\ 
 &    & R@0.5 & R@1 & R@2  \\ \hline
\multirow[t]{2}{*}{\textit{Single}}&Faster RCNN Resnet50 FPN Focal DIOU    &0.747	& 0.810 &	0.869  \\   
&Faster RCNN Swint FPN Focal DIOU  & 0.781 &	0.834  &	0.878  \\  \hline
\multirow[t]{2}{*}{\textit{Dual}} & \system\ (Resnet50) (ours)   &0.768	 & 0.840  &	0.878  \\
& \textbf{\system\ (Swint) (ours)}  & \textbf{0.797}  &	\textbf{0.852}  &	\textbf{0.895}  \\ \hline
\end{tabular}
\label{table:rvin}

\end{table*}

\subsection{Positional encoding}
To evaluate the impact of positional encoding, we trained our \system\ without positional encoding on VinDr-Mammo dataset. Table \ref{table:pos} shows that using positional encoding can improve the recall at every FPPI for both \system\ using SwinT and Resnet50 model. Specifically, at rate FFPI=1, the improvement is 2\% for \system\ SwinT and 3\% for \system\ Resnet50 model.
\begin{table}[ht]
\normalsize
\centering
\caption{This table shows the effect of positional encoding. We trained our multiview detector without positional encoding (no pos) and evaluated it on VinDr-Mammo dataset and compared it with the the model using VinDr-Mammo dataset}
\begin{tabular}{llll}
\hline
   &\multicolumn{3}{c}{Recall @ FPPI} \\ 
    & R@0.5 & R@1 & R@2  \\ \hline
\system\ (Resnet50) no pos  &0.768	& 0.810	 & 0.844  \\   
\system\  (Resnet50)   &0.768	 & 0.840  &	0.878  \\
\system\ (Swint) no pos   &0.788 &	0.831 &	0.873  \\
 \system\ (Swint)  & 0.797  &	0.852  &	0.895  \\ \hline
\end{tabular}
\label{table:pos}

\end{table}
\section{Cross transformer analysis} 
\label{sec:anal}
The intuition of our work is based on the assumption that the mass on two different views of the same breast will have relationships on shape, structure, and position which can be effectively encoded with cross-transformer. Moreover, we aim to demonstrate that leveraging these relationships can enhance our detector's performance, mirroring the practices of radiologists in their diagnostic workflow. We substantiate this intuition through two experiments.
\subsection{Does model decision based on two views?}

\begin{figure}[htbp]
    \centering
    \includegraphics[width=0.4\textwidth]{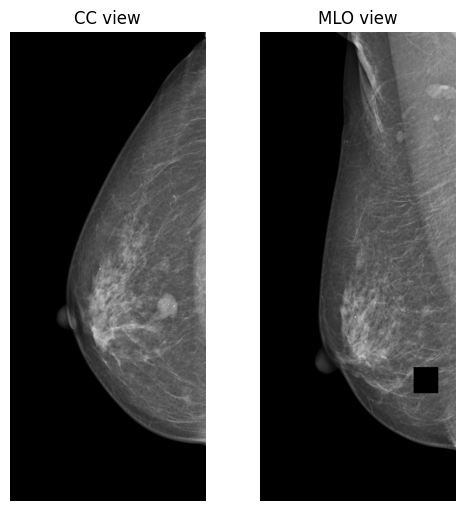}
    \caption{Synthesis dataset used for section \ref{sec:anal}. From VinDr-Mammo dataset, we masked out the bonding boxes for mass in MLO view while keeping the CC view be the same}
    \label{fig:synthesis}
\end{figure}
We created the synthesis dataset derived from VinDr-Mammo dataset where we masked out the masses in MLO view while keeping the CC view unchanged as illustrated in Fig \ref{fig:synthesis}. \\
To validate the model's ability to utilize information from both views effectively, we evaluate the performance of the multiview detectors using from both the synthesis dataset and the original VinDr-Mammo dataset. We report the FROC analysis on the CC view to see whether the information of the mass in one view (MLO view) can assist the model in diagnosing in the other view (CC view). Table \ref{table:mask} shows that the performance on the CC view drops significantly when the mass in the MLO view is masked out, aligning with our hypothesis.  
\begin{table}[htbp]
\normalsize
\centering
\caption{Evaluation on VinDr-mammo dataset for CC view only. We test our \system\ with Resnet 50 and Swint encoder with two different settings. The mask model is evaluated on VinDr-Mammo dataset but we mask out the lesions on MLO view and the other is evaluated on the original dataset.}
\begin{tabular}{llll}
\hline
   &\multicolumn{3}{c}{Recall @ FPPI} \\ 
    & R@0.5 & R@1 & R@2  \\ \hline
\system\ (Resnet50) mask  &0.675 &	0.789 &	0.825  \\   
\system\ (Resnet50)   &0.728	& 0.798	& 0.842  \\
\system\ (Swint) mask   &0.702	& 0.781	  &0.860  \\
 \system\ (Swint)  & 0.763	& 0.816& 0.886 \\ \hline
\end{tabular}
\label{table:mask}

\end{table}
\subsection{Cross-transformer as auto-register module}
\label{subsec:viz}

Registration between two ipsilateral views may provide useful information but it is a non-trivial task. Previous works \cite{perek}, \cite{YANyutong} employed auxiliary modules to align the finding between CC and MLO view to enhance model performance. Cross-transformer, in contrast, learns the relation of the finding between two views implicitly. The attention mechanism enables the model to assign higher weight to "important" RoIs, which, in this context, are the corresponding findings in the other view. In another view, cross-transformer serves as an auto-register module where it identifies the corresponding masses in the other view and assigns weight to combine it to generate the final result.  We also establish a quantitative evaluation of the registration ability of \system. Following the approach outlined by Yan et al. \cite{YANyutong}, we find the ipsilateral views in VinDr-Mammo dataset where each view has only one mass finding. We consider only the test portion of the dataset which resulted in 95 CC/MLO pairs, corresponding with $95$ masses finding in each view. In CC views, for each prediction with an Intersection over Union (IOU) with grouth truth box above 0.2, we calculate the relevance score introduced in Hilar et al \cite{Chefer_2021_ICCV} for this prediction and the RoIs in the corresponding MLO view. We select the RoIs with the highest score and check if they overlap with the growth truth box in MLO view. Table \ref{table:reg} shows the registration performance of \system\ using Swin Transformer. Our model can implicitly register with a recall and accuracy at $77.8\%$ and $77.9\%$ respectively, demonstrating that the model rely on the corresponding mass in another view to make the decision. It is worth noting that in our system, there are 1000 RoIs in each view, therefore, for the synthesis dataset, the accuracy and recall are zero. In Fig \ref{fig:viz}, the model correctly identifies the relevant ROIs from the original dataset, whereas it selects random ROIs in the synthesis dataset where the mass has been masked out.
\begin{table}[htbp]
\normalsize
\centering
\caption{Auto-registration performance of \system using SwinT as encoder. We evaluate on subset of VinDr-Mammo dataset describe in section \ref{subsec:viz} and the corresponding synthesis dataset.}
\begin{tabular}{lll}
\hline

    & recall & accuracy  \\ \hline
TransReg (SwinT) VinDr-Mammo &0.778 &	0.779   \\   
TransReg (SwinT) synthesis & 0	&  0 \\ \hline
\end{tabular}
\label{table:reg}

\end{table}


\section{Discussion and Conclussions} 
\label{sec:conclude}
In this paper, we utilize cross-transformer to enhance information fusion between ipsilateral views on mammograms. We evaluate the proposed method on two public datasets and archived SOTA FROC performance for mass detection problems. Furthermore, we also conduct experiments to provide evidence that the proposed method utilizes the information between two views effectively. The cross-transformer functions as an auto-registration module, replicating the diagnostic process employed by radiologists.\\
Due to limited resources, our models train on relatively small resolution compared with other work \cite{momminet}, \cite{high}. We also can only small batch sizes (2 or 4) and can not experiment with different hyper-parameter choices that affect model performance. The cross-transformer block only integrates information of region of interest of RCNN based detector in each view which is a late fusion mechanism. One possible solution is that we can embed the system to an object detector such as Detr \cite{detr} which create RoIs (object query) in very early stage. \\
In future work, we can leverage \system\ ability to handle other problems using multi-view such as chest Xray ( frontal and lateral views) or 3D object detection problem. We can also explore the system's ability to build an end-to-end breast cancer diagnostic system  and deploy it to hospital.
\bibliographystyle{IEEEtran}
\bibliography{LaTeX/cite}
\end{document}